\documentclass[12pt]{elsarticle} 
\usepackage[left=1.4cm,right=1.4cm,top=1.4cm,bottom=2.3cm]{geometry}
\usepackage{lineno,hyperref}
\usepackage{hyperref}  % 导入 hyperref 宏包
\usepackage{color}
\usepackage{subfigure}
\usepackage{fancyhdr}
\usepackage{epstopdf}
\usepackage{ulem}
\usepackage{bm}
\usepackage{amsopn}
\usepackage{amssymb,amsmath,color}
\usepackage{lineno}
\usepackage[algo2e]{algorithm2e} 
\usepackage{algorithm}

\biboptions{sort&compress}%参考文献

\begin{document}
%\modulolinenumbers[100]
%\linenumbers
\begin{frontmatter}
% %% Title, authors and addresses
\title{\Huge{\textbf{Support Vector Boosting Machine (SVBM): Enhancing Classification Performance with AdaBoost and Residual Connections}}}

% %% Group authors per affiliation: 
\author[1]{Junbo Jacob Lian}

\address[1]{School of Mathematics and Computer Sciences, Zhejiang Agriculture and Forestry University, Hangzhou 311300, PR China\\
(junbolian@qq.com)\\
*Corresponding Author: Junbo Jacob Lian (junbolian@qq.com)}

\begin{abstract}

In traditional boosting algorithms, the focus on misclassified training samples emphasizes their importance based on difficulty during the learning process. While using a standard Support Vector Machine (SVM) as a weak learner in an AdaBoost framework can enhance model performance by concentrating on error samples, this approach introduces significant challenges. Specifically, SVMs, characterized by their stability and robustness, may require destabilization to fit the boosting paradigm, which in turn can constrain performance due to reliance on the weighted results from preceding iterations. To address these challenges, we propose the Support Vector Boosting Machine (SVBM), which integrates a novel subsampling process with SVM algorithms and residual connection techniques. This method updates sample weights by considering both the current model's predictions and the outputs from prior rounds, allowing for effective sparsity control. The SVBM framework enhances the ability to form complex decision boundaries, thereby improving classification performance. The MATLAB source code for SVBM can be accessed at \href{https://github.com/junbolian/SVBM}{https://github.com/junbolian/SVBM}.

\end{abstract}


\end{frontmatter}
 
%% \linenumbers 

%% main text
\section{Introduction}

Boosting algorithms have long been recognized as powerful ensemble methods in machine learning \cite{1}, where weak learners are iteratively combined to form a more robust predictive model\cite{2}. In traditional boosting frameworks, particularly AdaBoost, the algorithm places higher importance on misclassified samples, enabling the model to focus on more challenging cases in subsequent iterations\cite{3}. This approach has been shown to significantly enhance classification accuracy across a wide range of tasks \cite{4,5,6}. As a result, AdaBoost has become widely adopted in the machine learning community, especially in applications involving integrated optimization with the tree model \cite{7,8,9}. Recently, the use of Support Vector Machines (SVMs) as weak learners within the AdaBoost framework has emerged as a novel direction in ensemble learning, offering promising applications across various fields. Y. Lan et al. proposed diagnostic algorithms for indirect bridge health monitoring using an optimized combination of AdaBoost and linear SVM \cite{10}. A. Belghit et al. optimized the One-vs-All SVM approach using the AdaBoost algorithm for rainfall classification and estimation from multispectral MSG data \cite{11}. L. Wen introduced a method for predicting the deformation behavior of concrete face rockfill dams by combining SVMs with the AdaBoost ensemble algorithm \cite{12}.

However, despite the strong performance of SVMs integrated with boosting in various domains, several challenges persist when employing SVMs as weak learners within the boosting process. SVMs are well-regarded for their stability, robustness, and excellent generalization capabilities, making them a popular choice for classification tasks \cite{13}. Yet, in a boosting framework, these advantages can become limitations. The inherent stability of SVMs conflicts with the iterative nature of boosting techniques, which require the model to dynamically adapt to misclassified samples. Consequently, traditional AdaBoost methods enhanced with SVMs may not achieve optimal performance \cite{14,15}. Additionally, SVMs are designed to be stable, making it difficult to “destabilize” them without compromising their overall performance. Their reliance on weighted sample data from previous iterations further restricts their ability to fully leverage the boosting mechanism.

Specifically, the output of AdaBoost is a linear combination of the outputs from its weak learners, and since each SVM learner's output is also a linear combination of its kernel outputs, the overall output can be simplified to a linear combination of these kernel outputs. This results in a unified support vector machine structure, which is a promising approach for designing monolithic SVM classifiers with improved performance derived from amplifying the benefits of the emphasis mechanism. However, the inherent stability of SVM classifiers—being linear in the approximate kernel Hilbert space—and their overall power suggests that directly using SVMs as RAB learners does not enhance the classification capabilities of the block design; in many cases, these capabilities may even diminish \cite{16}. Boosting design requires unstable and sufficiently weak learners to leverage the advantages of the step-by-step emphasis mechanism. If learners are stable, diversity does not manifest; conversely, if learners are too strong, the process converges in just a few iterations. Intentionally selecting "ad hoc" values for regularization parameters to weaken SVM classification units is not a viable alternative, as this approach tends to retain the same kernels (and hence the same Support Vectors), which reduces necessary diversity.

In recent years, researchers have explored various methods for boosting integrated designs with local approximation capabilities. For instance, some have employed indirect methods\cite{17}, introduced local gating mechanisms during the aggregation step of a global approximation learner \cite{18}, or developed Maximal Margin designs for large monolithic architectures by modifying other integration algorithms \cite{19}. While these integration models perform well, they often lead to high computational demands due to the independence between learners \cite{17,18} or the scenario where the number of elements significantly exceeds the number of training samples\cite{19}.

Conversely, several models have been proposed to destabilize and weaken SVM learners, falling into two primary categories. The first class of designs introduces diversity by varying kernel characteristics during the integration process \cite{14,20}. For example, the Diverse AdaBoost SVM (DABSVM) \cite{14} uses a traditional Gaussian kernel and requires each new learner to exhibit diversity relative to previous ones. However, this approach does not successfully reduce the integration to a singleton equivalent form. The second class of improved SVM boosting integrations utilizes subsampling methods to obtain appropriate SVM learners \cite{21}. One such approach, AdaBoost Weakness SV (ABWSV)\cite{22}, employs a v-SVM design to better manage the characteristics of the base classifiers \cite{23}.

Despite these two strategies, success in destabilizing and weakening SVM learners has been limited. To tackle these challenges, E. Mayhua-López et al. recently proposed a more effective improvement technique that combines two key components \cite{24}. Firstly, they devised a sophisticated structured subsampling mechanism that avoids the repetitive selection of difficult samples, thereby preserving diversity while weakening the learner—contrasting with previous subsampling mechanisms aimed solely at enforcing diversity among integrated units. Second, they reconfigured the sparsity-inducing SVM training algorithm LPSVM \cite{3,17,18,19,20} by employing a kernel matrix with row subsampling, integrating selected samples as kernels while training all available samples. Not only does LPSVM effectively weaken the learner, but the bi-domain linear programming (LP) also helps control the number of kernels included in the SVM solution, facilitating the construction of a compact and equivalent monolithic final architecture. However, this model still relies heavily on weighted sample data from prior iterations, which can limit its effectiveness in fully utilizing the boosting framework.

To address these challenges, we propose the Support Vector Boosting Machine (SVBM), which integrates a subsampling strategy and residual connection techniques within the boosting process. By incorporating subsampling, we enable SVMs to be trained on diverse subsets of data, helping to mitigate overfitting and enhancing model flexibility. The residual connection method draws inspiration from the residual learning technique (ResNet) introduced by K. He et al. for convolutional neural networks \cite{25}. In SVBM, this residual linking mechanism ensures that weight updates do not solely depend on the predictions of the current iteration; instead, they also incorporate information from previous rounds. This integration of past and present predictions allows for more nuanced weight adjustments, effectively controlling sparsity and enabling the model to establish more complex decision boundaries. The main contributions of this study are as follows:

1.	\textbf{Proposed SVBM Framework:} We introduce the SVBM classification framework, which enhances the adaptability and effectiveness of SVM-based weak learners through the integration of subsampling and residual connections.

2.	\textbf{Comprehensive Experimental Validation:} Through extensive experiments conducted on ten publicly available datasets, we demonstrate that SVBM outperforms existing boosted SVM models in terms of classification performance and generalization ability. This is supported by a thorough comparative analysis, including ablation studies.

3.	\textbf{Open-Source Accessibility:} We provide the open-source code and framework for SVBM, offering an easy-to-use tool that researchers can readily apply across various fields. This accessibility promotes further exploration and application of the proposed method in real-world scenarios.

The paper is structured as follows:  \textbf{Section 2} provides an in-depth exploration of the architecture and specific details of the SVBM model. \textbf{Section 3} presents the model ablation and comparative experiments conducted to validate SVBM's design, showcasing its performance on ten well-known public datasets. \textbf{Section 4} summarizes the findings of this study and offers insights into potential avenues for future research.
\section{Support Vector Boosting Machine (SVBM)}
\subsection{SVM model}
SVM is a supervised learning model commonly employed for binary classification tasks, aiming to maximize the inter-class margin by identifying an optimal hyperplane that separates samples from different classes as distinctly as possible. To address multi-classification tasks, we utilize an extension of the multiclass SVM model known as the Error-Correcting Output Codes (ECOC) method. This approach combines multiple binary classifiers to create a classification model capable of effectively handling multiclass problems.

The Radial Basis Function (RBF) kernel is one of the most widely used kernel functions in SVM, particularly suited for tackling nonlinear and non-differentiable problems. In our model, we incorporate the SVM based on the RBF kernel as a weak learner within the SVBM framework, alongside the ECOC method. This combination allows us to decompose the multi-classification task into several binary classification problems by constructing multiple binary classifiers and generating codes for different categories. Each binary classifier addresses a specific classification problem between two categories, and the final classification decision is determined by the voting or weighted results of these multiple classifiers.

SVM with RBF kernels can be enhanced through intelligent optimization algorithms, which represent a significant improvement for the Support Vector Boosting Machine (SVBM) \cite{26,27,28}. Additionally, optimizing SVBM using Linear Programming Support Vector Machines (LPSVM) could also serve as a potentially effective approach.

\subsection{Subsampling AdaBoost model based on SVM}
In pursuit of enhancing the classification accuracy of the SVM model, this approach introduces a pioneering strategy that leverages the Boosting technique to reinforce data learning in the presence of judgment error (depicted in Fig. 1). This methodology employs a weak learning machine (WLM), specifically the SVM, in a multi-step framework. The Refs. \cite{29,30} provide further details. It is important to highlight that the AdaBoost model employed in this experiment incorporates a subsampling selection mechanism. To ensure diversity and fairness in each training round, samples are not simply drawn from the instances with the highest weights; instead, structured sampling is performed across the entire dataset in proportion to the weights. Specifically, a sampling proportion   is defined, and the top   samples with the largest weights are selected from the sorted sample set in each round. The value of the sampling ratio   is determined by the dataset size and the training requirements of the model, typically set to half the size of the training set. The ensuing process encompasses the following approximate steps:

\begin{figure}[H]%%[H]可以强制图表位置
	\centering
    	\includegraphics[width=16cm,height=7cm]{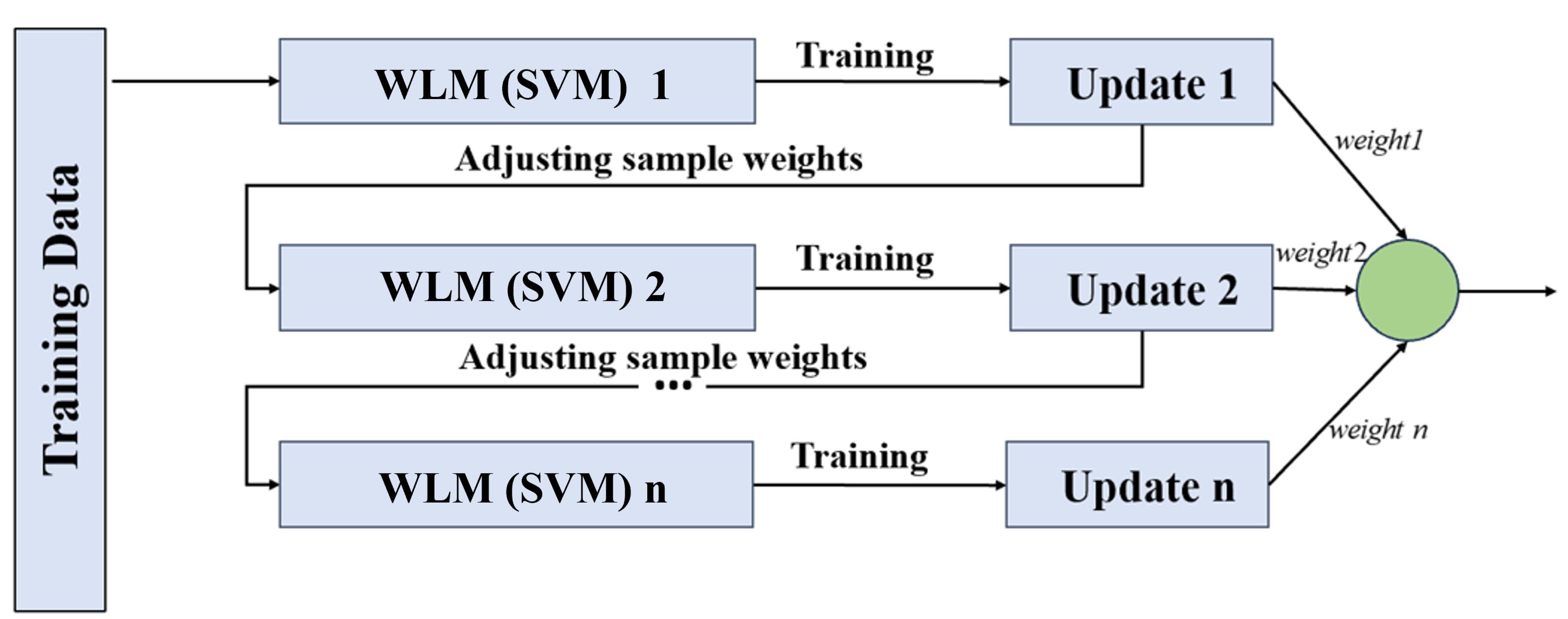}
	\caption{AdaBoost-SVM model}% 给图片加title，可以自动编号
	\label{fig1} %给图片加label用于超链接
\end{figure}

1.	\textbf{Initialization of Sample Weights:} The weights of each sample in the training set (consisting of   samples) are initialized to 1. These weights indicate the difficulty level of correctly classifying each sample. The model learning rate, or weight growth rate, is set to 1.

2.	\textbf{Iterative Training: }  iterations are performed, where each iteration involves training a WLM using the current weighted training set.

3.	\textbf{Weak Learner Training: }A SVM is chosen as the WLM and trained using the weighted training set. The objective of the WLM is to minimize the weighted error rate, considering the difficulty level of each sample as reflected by their weights.

4.	\textbf{Evaluation of the WLM: }The weighted error rate of the WLM is calculated by examining the predictions made by the WLM compared to the true labels. The weighted error rate is computed using the following formula:

\begin{equation}
   \varepsilon=\frac{\sum_{i=1}^{N} w_{i} \cdot \mathbb{I}
   \left(h\left(x_{i}\right) \neq y_{i}\right)}{\sum_{i=1}^{N} w_{i}}
    \label{eq1}
\end{equation}

Here, $N$ represents the number of samples in the training set, $w_i$denotes the weight of the $i^{th}$ sample, and  $\mathbb{I}(h\left(x_{i}\right) \neq y_{i})$ is an indicator function. The indicator function equals \ref{eq1} when the prediction $h(x_i)$ of the $i^{th}$ sample is not equal to the true label $y_{i}$, and 0 otherwise. The weighted error rate provides an assessment of the WLM's performance on the current weighted training set.

5.	Calculation of Classifier Weights: The weights of the WLMs are calculated based on their weighted error rates. Classifiers that exhibit higher accuracy are assigned higher weights. The weight of a WLM (classifier weight) is computed using the formula:

\begin{equation}
   \alpha=\frac{1}{2} \ln \left(\frac{1-\varepsilon}{\varepsilon}\right)
    \label{eq2}
\end{equation}

where $\varepsilon$ represents the weighted error rate and   denotes the weight of the WLM.

6.	Update of Sample Weights: The weights of the samples are adjusted according to their classification results obtained from the WLM. The weights of misclassified samples are increased to emphasize their importance in subsequent iterations, while the weights of correctly classified samples are decreased to reduce their influence.

7.	Normalization of Sample Weights: The sample weights are normalized to ensure their sum equals \ref{eq1}. This normalization step preserves the interpretation of the weights as probabilities.

8.	Repeat Steps 2-7: The training process is repeated for a specified number of iterations or until the desired level of performance is achieved. In each iteration, the steps of WLM selection, training, evaluation, classifier weight calculation, and sample weight update are repeated.

9.	Combination of WLMs: The trained WLMs are combined to form a strong learning machine by assigning weights to each WLM. The weights of the WLMs are determined based on their performance in the Boosting process.

\subsection{Residual connection}
In this study, we introduce the concept of residual connections to enhance the performance of our boosting learning algorithm. Residual connections integrate information from both current and previous states, allowing the model to better handle complex data patterns. In our implementation, the residual connection is primarily manifested in the weight update mechanism for the training samples.

Initially, each training sample is assigned equal initial weights, ensuring that the model treats all samples equally during the first iteration. As the algorithm progresses, the weights are updated based on the predictions made by the current classifier. The update formula is expressed as follows:

\begin{equation}
   \text { weights }=\text { weights } \times \exp \left(-a \times\left(2 \times\left(Y_{\text {pred }}==Y_{\text {train }}\right)-1\right)\right) 
    \label{eq3}
\end{equation}

Here, $\alpha$ represents the weight of the current classifier (see Eq. (\ref{eq2})).

The term $error_t$ denotes the weighted error rate of the current classifier. To avoid computational issues with logarithmic calculations, we set $error_t$ to 0.4999 if it exceeds 0.5.

The introduction of residual linkage in the weight updating method may result in the current round's weights being influenced by those of the previous round. If the model's predictions in the previous round perform poorly, this influence can negatively affect the current weight update, potentially leading to a sudden drop in accuracy. This effect is particularly pronounced in the early stages of training, where poor model performance in initial rounds can propagate undesirable outcomes to subsequent iterations through residual connectivity. To mitigate this issue, it is necessary to strike a proper balance between the current predictions and the previous round's weights when updating the weights. To achieve this, the model incorporates a parameter, $\beta$, to control the impact of the residuals.

After updating the weights based on the classifier's performance, we introduce a residual connection by combining the current weights with the previous round's weights. This is achieved through the following equation:

\begin{equation}
    weights=\frac{weights+b \times prev\_weights}{1+b}
    \label{eq4}
\end{equation}

In this formulation,$\beta$ serves as a hyperparameter that controls the influence of previous weights on the current weights. By storing the weights from the previous training iteration in $prev_weights$, we ensure that the model retains valuable historical information during weight updates. This approach enhances robustness against sample imbalance and noise, which are common challenges in machine learning.

To maintain the integrity of the weight distribution, the weights are normalized after each update, ensuring that their sum equals one. This normalization step preserves the relative relationships among weights, enabling the model to effectively focus on different samples in each training round.

We further optimize the hyperparameter $\beta$ through a dynamic adjustment strategy. Initially set to 0.5, $\beta$ can adapt based on the cumulative accuracy of the classifier. If the cumulative accuracy of the current round is lower than that of the previous round, we slightly increase $\beta$ to enhance the influence of the previous weights:
\begin{equation}
   \beta=\min \left(1, \beta \cdot 1.05 \cdot\left(1-\frac{t}{2 \cdot n_{classifiers }}\right)\right)
    \label{eq5}
\end{equation}

Conversely, if the cumulative accuracy improves, we decrease   to reduce reliance on previous weights:
\begin{equation}
   \beta=\max \left(0, \beta \cdot 0.95 \cdot\left(1-\frac{t}{2 \cdot n_{\text {classifiers }}}\right)\right)
    \label{eq6}
\end{equation}

The beta parameter in the design is integrated into the adaptive fluctuation descent function. This adjustment is crucial because, in the early stages of training, the model's understanding of the data is relatively shallow, and the classifier's performance may not be optimal. At this stage, a higher $\beta$ value should be used to fully leverage the previous weights, promoting faster model learning. A higher $\beta$ value enables the model to absorb knowledge from earlier rounds more effectively, accelerating the improvement of the current model. Additionally, this higher beta value allows the model to respond more quickly to misclassified samples, helping it to adapt to the data distribution in the initial rounds and improving the convergence speed of training.

However, as the model matures, relying too heavily on a high $\beta$ value may cause the model to over-depend on previous predictions, particularly when the model has stabilized in the later stages. At this point, the $\beta$ value should be reduced to decrease the reliance on historical weights, granting the model greater flexibility to adjust its own weights. Lowering the $\beta$ value in the later stages helps reduce the risk of overfitting, as the model has already learned substantial characteristics of the data. This adjustment allows the model to balance new data with its current weights more effectively, ensuring that it can capture new patterns and trends as training progresses.

This adaptive mechanism allows $\beta$ to flexibly respond to changes in model performance and sample characteristics throughout the training process. By balancing the influence of current and historical weights, we enhance classification accuracy and model robustness.

The integration of residual connections and adaptive $\beta$ adjustments significantly improve the algorithm's performance, as evidenced by the enhanced classification accuracy and stability observed in our experiments on the dataset.

\subsection{Pseudo-code of the SVBM}
The SVBM model integrates SVM, a subsampling technique, and a residual connection mechanism to enhance classification accuracy. The core innovation lies in the weighted averaging of sample weights from the current and previous rounds via residual linkage, which improves the stability of the weight update process. Additionally, the model dynamically tunes the beta parameter to adapt to different stages of training. The pseudo-code for this approach is presented in Algorithm 1.

\begin{algorithm}[H]
\label{algorithm1}
  \SetAlgoLined
Initialize weights, alpha, and parameters\;

\For{i = 1:n\_classifiers}{
     Select subset of data using structured\_subsampling\;
    \\
    Train SVM on subsampled data\;
    \\
    Predict on training data\;
    \\
    Compute error and update alpha\;
    \\
    Update weights using residual connection\;
    \\
    Adjust beta based on accuracy progression\;
    \\
    Visualize weight distribution and accuracy progression\;
 }\leavevmode 
\textbf{For} each classifier, predict on test data
\\
Combine test predictions and compute final accuracy
\\
 Visualize confusion matrix and test classification
\caption{: SVBM model}
\end{algorithm}

\end{document}